\documentclass{article}

   \usepackage[preprint]{neurips_2020}

\usepackage[utf8]{inputenc} 
\usepackage{authblk}
\usepackage[T1]{fontenc}    
\usepackage{hyperref}       
\usepackage{url}            
\usepackage{booktabs}       
\usepackage{amsfonts}       
\usepackage{nicefrac}       
\usepackage{microtype}      
\usepackage{graphicx}

\usepackage{epstopdf}

\title{\textit{FathomNet}: An underwater image training database for ocean exploration and discovery}
\author[1]{Océane E. Boulais*\thanks{\texttt{oceane@mit.edu}}}
\author[2]{Ben Woodward*}
\author[3]{Brian Schlining}
\author[3]{Lonny Lundsten}
\author[3]{Kevin Barnard}
\author[1]{Katy Croff Bell}
\author[3]{Kakani Katija}
\affil[1]{MIT Media Lab}
\affil[2]{CVisionAI}
\affil[3]{Monterey Bay Aquarium Research Institute}

\begin{document}
\maketitle
\begin{abstract}
 Thousands of hours of marine video data are collected annually from remotely operated vehicles (ROVs) and other underwater assets. However, current manual methods of analysis impede the full utilization of collected data for real time algorithms for ROV and large biodiversity analyses. \textit{FathomNet} is a novel baseline image training set, optimized to accelerate development of modern, intelligent, and automated analysis of underwater imagery. Our seed data set consists of an expertly annotated and continuously maintained database with more than 26,000 hours of videotape, 6.8 million annotations, and 4,349 terms in the knowledge base. \textit{FathomNet} leverages this data set by providing imagery, localizations, and class labels of underwater concepts in order to enable machine learning algorithm development. To date, there are more than 80,000 images and 106,000 localizations for 233 different classes, including midwater and benthic organisms. Our experiments consisted of training various deep learning algorithms with approaches to address weakly supervised localization, image labeling, object detection and classification which prove to be promising. While we find quality results on prediction for this new dataset, our results indicate that we are ultimately in need of a larger data set for ocean exploration.

\end{abstract}

\section{Introduction}

Today, we gather thousands of hours of underwater imagery and video using Remotely Operated Vehicles (ROVs) and other underwater assets, but remain unable to mobilize its full value potential because of the prohibitive amount of time that is required to manually review the video data, and is subsequently not viewed. While this video data is collected by a wide variety of research groups in the marine sciences across the globe, the analysis of the data collected cannot be fully realized until there exists an exemplar annotated repository of marine life imagery. In this paper we propose \textit{FathomNet}: a baseline image training set that is optimized to accelerate development of modern, intelligent, automated analysis of underwater imagery. In addition to the data set, this paper describes the exploratory experiments with deep learning algorithms that were used to create a baseline architecture for future work. Applications for this body of research span a wide spectrum, from accelerated video review and real time algorithms for ROVs, to large biodiversity analyses. Creating application-primed data sets has been an effective tactic to increase research accessibility and model robustness. Recent work on characterizing large data sets has demonstrated the importance of representative data for targeted applications. Attributes such as background, image viewpoint, and scale have dramatic effects on the generalization of algorithms to real-world scenarios.
\par We began with focusing on weakly supervised localization because we sought an algorithm that could propose bounding boxes that could be manually corrected and verified by human experts. This is a workflow that is known to accelerate annotation time, and reduce the strain of producing so many annotations for a user. The structure of the seed data set was such that most of the frame grabs had image level labels associated with them.  The ultimate goal of the \textit{FathomNet} data set was to enable modern Convolutional Neural Network (CNN)-based object detection and classification algorithms to be developed for species that existed in seed data sets, crucial in demonstrating the viability and potential of the data set.

\section{Technical Approach}
\label{gen_inst}

With the advent of modern deep learning architectures and the abundance of annotated data sets, the importance of representative data for targeted applications has been effectively demonstrated. Niche data sets such as \textit{ObjectNet} (Barbu, A. et al., 2019) and Fashion-MNIST (Zalando, 2017) have built upon the generalizeable properties of MNIST (LeCun et al., 1998) and \textit{ImageNet} (Deng et al., 2009). Until now, a baseline training data set for the analysis of underwater imagery has been unavailable. Using algorithms to accurately detect regions of interest (ROIs) in non-iconic imagery is critical for the progress of automated inspection of video that will be collected on expeditions for years to come. We introduce the provenance of the seed data in Section 2.1, and in Section 2.2 we describe the data set and their respective annotations. Full access to \textit{FathomNet} and existing models are available in the Supplemental Material section.

\begin{figure}[h]
    \centering
    \includegraphics[width=\textwidth]{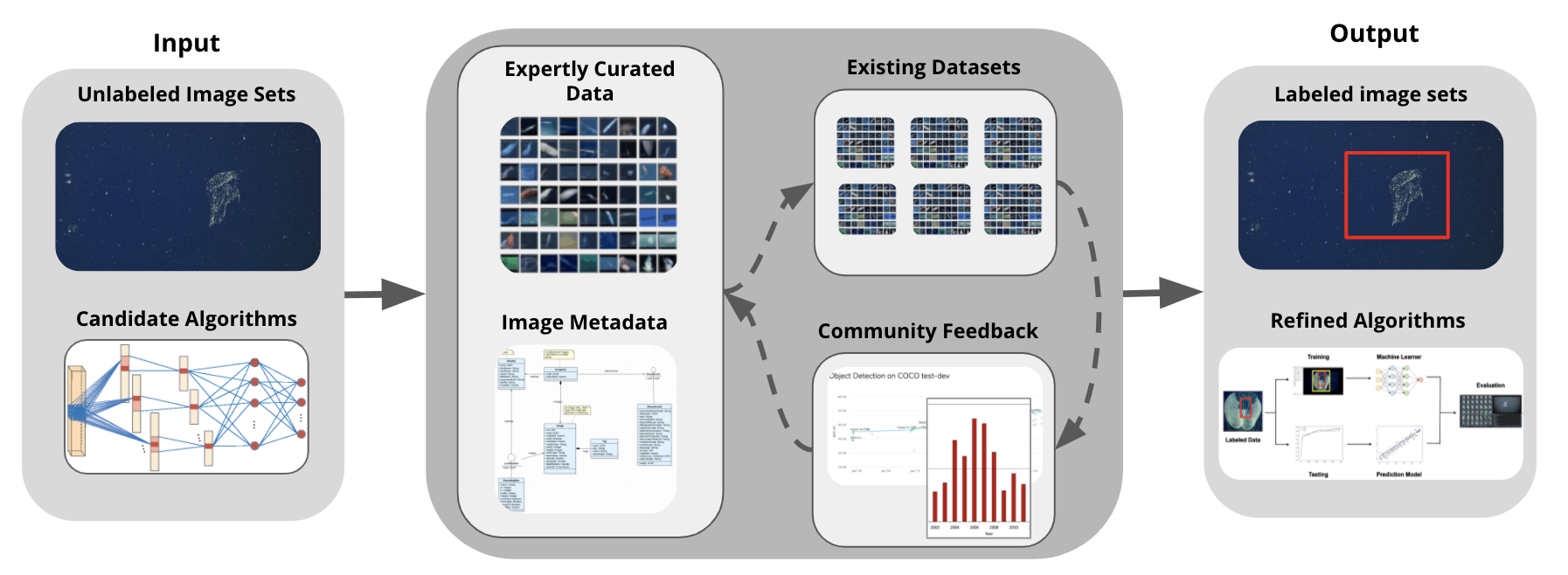}
    \caption{An overview of our efforts to develop \textit{FathomNet}.}
    \label{fig:overview}
\end{figure}

\subsection{Seed data set}

The seed data set used high-resolution video equipment to record hundreds of remotely and autonomously operated vehicle dives each year. This video library contains detailed footage of the biological, chemical, geological, and physical aspects of each deployment. Our data set consists of more than 26,000 hours of videotape that has been archived, annotated, and maintained as a centralized resource. This resource is enabled by the Video Annotation and Reference System (VARS; Schlining and Jacobsen Stout 2006), which is a software interface and database system that provides tools for describing, cataloguing, retrieving, and viewing the visual, descriptive, and quantitative data associated with deep-sea video archives. All of the video resources are expertly annotated by members our team, and there are currently more than 6.8 million annotations and 4,349 concepts (or classes) in the VARS knowledge base, with over 2000 of those concepts belonging to either genera or species.  
\par Using the VARS Query tool, a list of annotations that describe different genera and geologic features can be obtained. Of those annotations, each genus (of midwater and benthic animals) and geologic feature were ranked by the number of associated frame grabs. Drawing on the nomenclature used in the Panoptic Segmentation (Kirilov, He, Girshick, Rother,\ \& Dollar, 2018) task, these correspond to “stuff” and “things”. Upon selecting the top 18 midwater genera, the top 17 benthic genera, and the top 3 geologic features to incorporate into \textit{FathomNet} (Figure \ref{fig:overview}), the initial phases of this effort were to automate the classification of “things” (or animals) instead of “stuff” (or geological features), and the image set was then divided into Midwater and Benthic classes. 

\begin{figure}[h]
    \centering
    \includegraphics[width=\textwidth]{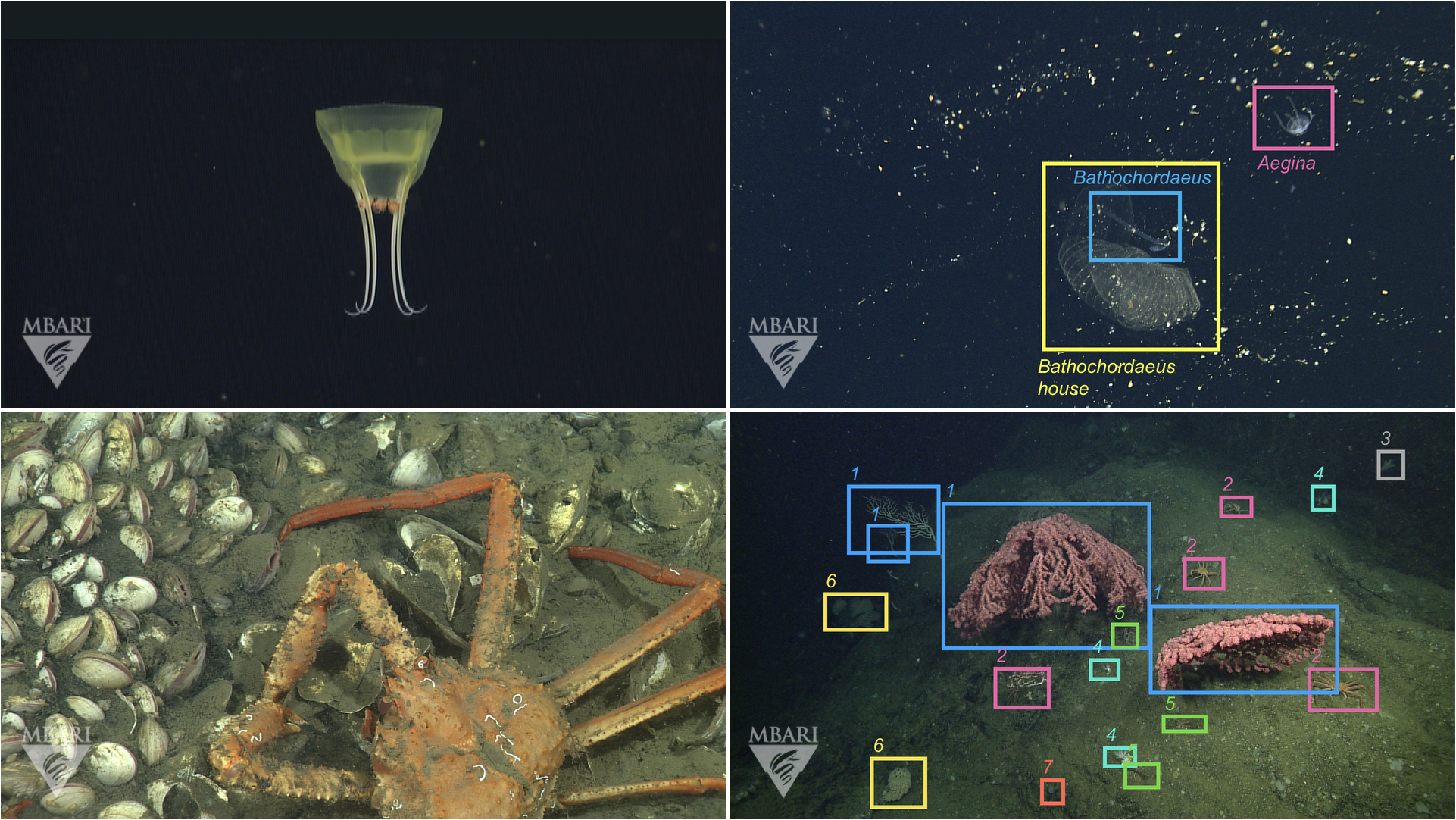}
    \caption{Example of the disparity between the available annotated data in the oceanographic community versus the level of annotations that are required for the data science community. While the top and bottom images in the left column were originally assigned single-concept annotations for jellyfish \textit{Aegina} and crab \textit{Chianoecetes}, respectively, they lacked localization data. The top and bottom images in the right column were originally assigned single-concept annotations for \textit{Aegina} and \textit{Chionoecetes}, respectively, and again exclude localization data. Although the images on the left are largely single-concept, the images on the right are obviously multi-concept. Our effort, as described here, was to resolve these disparities between research communities by creating \textit{FathomNet}, augmenting oceanographic annotations by adding multiple concepts in an image and adding localization data, as demonstrated by the localized images on the right. The top and bottom rows show representative images from midwater and benthic environments, respectively. The left and right columns show examples of iconic and non-iconic imagery, respectively. (Top-Left) Iconic images of \textit{Aegina} and (Bottom-Left) \textit{Chionoecetes}. (Top-Right) Non-iconic image of midwater concepts that include \textit{Aegina}, a larvacean \textit{Bathochordaeus}, and a \textit{Bathochordaeus} house. (Bottom-Right) Non-iconic image of benthic concepts that include (1) \textit{Paragorgia} coral, (2) \textit{Chianoecetes}, (3) \textit{Asteroidea}, (4) \textit{Psolus}, (5) \textit{Pandalopsis}, (6) \textit{Heterochone}, and (7) \textit{Tunicata}.}
    \label{fig:iconic}
\end{figure}

\subsection{Data set description: midwater and benthic classes}
The ocean is divided into different zones based on a variety of characteristics that include depth and light. The surface waters of the ocean are commonly described as the region where light penetrates, and generally extends from the surface to 200 m deep. The benthic region extends from the bottom of the ocean to 50 m above it, and the midwater region is the habitat that connects the surface waters and the benthic region. The midwater and benthic regions together make up the largest habitable ecosystem and is arguably the least explored on the planet. While animals that live in the benthic region are often associated with a variety of substrates, midwater animals inhabit a region effectively without boundaries, and has implications for the types of imagery that are collected in both places. These visual differences between habitats necessitates separating midwater and benthic classes for our subsequent efforts. The frame grabs and labels for the top midwater concepts include 18 midwater genera, with a mix of iconic and non-iconic views. The benthic imagery consisted of 17 classes, where each described a benthic animal genus. The nature of the frame grabs and the distribution of species meant that images often only contained a single species of interest, which often corresponded to the label that came along with the image.

\subsection{Evolution of data annotation state}
 The data science and oceanographic communities use the term "annotation" differently. The seed data set was expertly curated, and "annotated", but the annotations lacked localization data for the object(s) of interest. Further, when annotations were associated with a specific video frame or image, those images were not exhaustively annotated. In many cases, a frame grab would correspond to an annotation (i.e., single-label) despite many instances where there were more objects of interest of the same or different classes within the image. For operational purposes, algorithms were desired that could identify all objects of interest within a scene, also known as multi-label image classification. Single label annotations provide training data that is severely mismatched to the desired algorithm outputs (Zhao, Zu \& Wu, 2019). The majority of our initial data being singly-labeled while being strongly multi-label in reality meant that our early algorithmic experiments revolved around using noisy labels to help identify ways to augment the data. This approach had a limited degree of success, and soon informed improvements to annotations which included exhaustively annotating multiple labels with bounding boxes within single-label imagery. Based on these recommendations, we began generating localized, multi-label data, and are focused on improving this process going forward.
\par Finally, we have started to incorporate imagery with other characteristics, to help improve the diversity of the data set. For example, many of the initial frame grabs were from cropped or zoomed-in iconic views of objects, and these views are often vastly different from the desired operational use cases for algorithms developed with this data set. Therefore we have begun including data of non-iconic imagery that were collected in both midwater and benthic environments.

\section{Algorithm Experiments}
\label{gen_inst}
In addition to enriching the data with localization annotations, we experimented with three different classes of algorithms with the desire to accelerate data set generation. As described in Section 2.2, the initial state of the data included significant label noise at the image level due to the single-label assignment in strongly multi-label images. These algorithms included image classification, weakly supervised localization, and object detection.

\subsection{Image Classification - ResNet 50}
\begin{figure}[h]
    \centering
    \includegraphics[width=\textwidth]{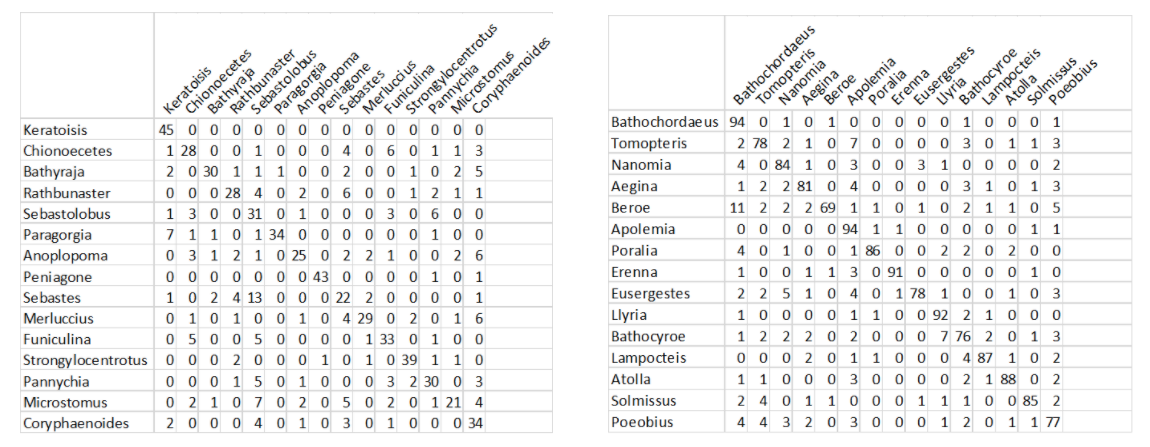}
    \caption{Confusion matrices for benthic (left) and midwater (right) algorithms}
    \label{fig:matrix}
\end{figure}

We trained a basic image classification algorithm on the benthic and midwater data sets using an \textit{ImageNet} pre-trained ResNet50 architecture. To account for the strong multi-label nature of the data we utilized both a Top-1 and Top-3 scoring metric, where the results can be seen in Figure \ref{fig:matrix}. While there are ways to train networks, such as ResNet, for multi-label problems (Gardner, Nichols, 2017;Wang, Jia, Breckon, 2018;Li, Yeh, 2018;Wang et al., 2017), our initial data set only provided single-label imagery. For both subsets of data, 80\% was used for training, 10\% for validation, and 10\% for testing. We report performance on the test split. Further, to test the ability of the algorithms to work in an operational setting, we ran the algorithm on data from video transects. During a transect, an underwater vehicle transits at a consistent speed for a consistent duration using the same imaging field-of-view at different observational depths. Unlike discovery modes, where an underwater vehicle is piloted to search for and observe animals in close quarters, transect data involves observations of animals or targets of interest that move steadily past. These transects are not balanced across classes, and provide markedly different views of the data, particularly for the midwater transects. 

\par For the midwater data, a classifier was trained for 15 species, which were filtered based on the number of images for each genus, and with a cutoff of approximately 1,000 images per genus. This resulted in a data set of 33,064 images. The classifier was trained for the concepts using only the single-class labels, and was able to identify multiple concepts within an image using a Top-N scoring methodology.
\par A rigorous application of this technique would produce false positives for each of the Top N concepts, however the relative accuracy and occurrence of misidentified classes was demonstrated as part of this process. Figure \ref{fig:gradcam} shows the resulting confusion matrix for the \textit{FathomNet} data set. Using the aforementioned scoring methodology, we found the Top 1 and Top 3 accuracies to be 85.7\ \% and 92.9\ \%, respectively on the midwater test set.

\par We then evaluated algorithm performance by on a midwater video transect data set. For many midwater taxa of interest, the \textit{FathomNet} training set consisted of a fair amount of zoomed-in, iconic views of animals. These views contrasted sharply with the limited spatial resolution of targets in the midwater transect footage, as these animals passed quickly by the underwater vehicle at a relatively far distance. Therefore, due to differences in imagery between the midwater transect data (e.g., differences in scale and resolution of objects) and the \textit{FathomNet} training data (e.g., iconic imagery), we found that the model performed poorly, and is consistent with work characterizing model generalization difficulties around parameters such as scale and viewpoint (Barbu, A. et al., 2019; Oksuz et al., 2020). The difference in spatial scales was akin to showing someone iconic pictures of sports cars, and asking them to identify these vehicles from aerial photos.

 \par We also trained an algorithm on 15 of the 17 benthic classes, initially setting the minimum number of images for the training set at 700 and later removing this limit to include classes that were abundant in the video transect data. This resulted in a data set with a total of 33,064 images.  Top 1 and Top 3 accuracies of 72.4\ \% and 92.8\ \%, respectively were achieved for the test set. An immediately noticeable difference between the benthic and midwater results is the difference in Top 1 vs Top 3 accuracy metrics. There is a marked improvement in the benthic data set moving to Top 3 accuracy. The reason for this is due to the strongly multi-concept nature of benthic imagery in \textit{FathomNet}, while the midwater images tended to be mostly single-concept. In many cases the label assigned to a benthic image was not the dominant concept from the data set within the image. This created a number of challenges, but also opened up interesting research areas in multi-label image algorithms and noisy data sets. An example of a multi-concept image is shown in Figure \ref{fig:gradcam}. Therefore, the remainder of our efforts and  discussion on algorithm development focuses on the benthic imagery because of the more appropriate nature of the test transect video, as well as the similar nature of video and images to other collaborators’ data. 

\subsection{Weakly Supervised Localization - GradCAM++}
To assess our ability to generate an object proposal algorithm from the single-label imagery, we explored weakly supervised localization techniques (Gao, Li, Yu, Morariu, \& Davis, 2018;Najibi, Yang, Wang, \& Piramuthu, 2018;Papadopoulos, Uijlings, Keller, \& Ferrari, 2016). Specifically we used GradCAM++ (Chattopadhyay and Sarkar, 2018) on the ResNet50 classifier from the previous section. We explored two ways to generate saliency maps and bounding box proposals: class-specific search and dominant class search. Dominant class search uses the class label from the output of the classifier to inform which concept to search for in an image, while class specific search selects a specific label to search for. Figure \ref{fig:hierarchy} shows how this technique nominally work, with the output saliency map being generated for the dominant class label from the classifier.
\par The results of the experiments with this algorithm showed that they could be useful under certain fairly restrictive conditions. In general, they worked best when the class being searched for was the most prominent class of interest in the image. Additionally, it worked well at localizing up to a few instances of a class, but started to degrade in effectiveness when increasing past that number. Recent works have proposed limitations to CAM methods for visual explanation (citation Neurips 2018 paper), or improvements (Ablation CAM). From an operational perspective, we note that in order to identify localize multiple concepts within an image, the technique would have to be run using class specific search for every class of interest, greatly increasing computation time. While not unfeasible, we identified other areas of effort as higher priority, and left these investigations to future efforts.
\par
\begin{figure}[h]
    \centering
    \includegraphics[width=\textwidth]{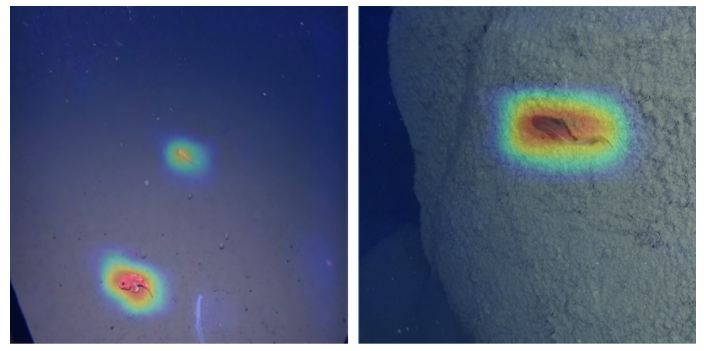}
    \caption{GradCAM++ results from (left) NOAA’s ROV \textit{Deep Discoverer} and  (right) National Geographic Society’s DropCam respectively. For both images, the model used was a 12 class ResNet50 trained on MBARI’s benthic data. For the Okeanos clip (left), the model generated a saliency map for the class Sebastolobus, correctly detecting multiple Sebastolobus in the footage. For the DropCam clip (right), the model generated a saliency map for the class Merluccius, detecting a similar-looking class of animals in the footage.}
    \label{fig:gradcam}
\end{figure}

 In addition to our algorithm experiments on the seed data footage, we also obtained video from National Oceanic and Atmospheric Administration's (NOAA) ROV \textit{Deep Discoverer}, as well as National Geographic Society’s \textit{DropCam}. We applied GradCAM++ from the ResNet50 trained classifier on videos from each of these external sources, and obtained very promising results (Figure \ref{fig:gradcam}). 

\subsection{Object Detection - RetinaNet}
The final class of algorithms that we experimented with were object detection and classification techniques. Specifically we used the RetinaNet (Lin, Goyal, Girshick, He, \& Dollar, 2017) single-stage algorithm with a ResNet50 backbone. In order to use this algorithm, the The seed data set Video Lab exhaustively annotated over 3000 benthic images with bounding box information for over 200 species. This resulted in approximately 23,000 bounding box annotations.  As with most data sets of this type, we suffered from the long tail problem, with the large majority of  annotations belonging to only a handful of classes. 
\par
\begin{figure}[h]
    \centering
    \includegraphics[width=80mm]{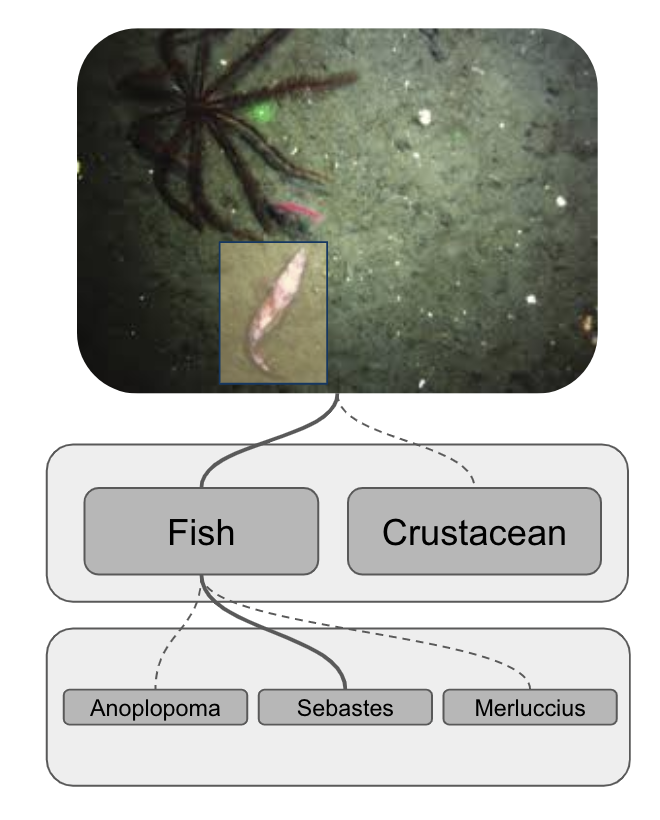}
    \caption{Example hierarchy showing how a multi-tiered algorithm could be used. Current algorithm efforts will help annotators efficiently label concepts to the finer hierarchical level , and eventually a multi-stage algorithm could be developed to conduct the entire hierarchical classification.}
    \label{fig:hierarchy}
\end{figure}

 We considered two approaches to overcome this limitation: 1) Collapse all labels to an “object” category and train a strong candidate object detector for classification by a human expert, and 2) Create a hierarchy for the labels and try to operate at limited taxonomic resolution. We experimented with the second approach to see how we might utilize combinations of algorithms, non-expert humans, and expert humans in the review process. Figure\ref{fig:hierarchy} shows an example of how this workflow could work. Two promising hierarchies were identified, and included a high-level concept (e.g. fish and crustacean), and morphological appearance (e.g., fish-like, laterally flattened, fan-like, and plant-like).
\par Our results show that we were able to create a strong object of interest detector, but that the automated labels did not provide additional efficiency compared to the existing expert annotators correcting and labeling identified objects of interest. Identifying different hierarchies and potential semi-automated workflows is an ongoing effort.

\section{Discussion and Future Steps}
The ability for algorithms to work under data set shift and out of distribution (OOD) data has been explored recently in the literature such as \textit{ObjectNet} and data set Shift (Ovadia, et al, 2019). Typically, explorations have either centered on the ability of the algorithm to generalize under data set shift or detect OOD data. Alternatively, the \textit{ObjectNet} data set showed that inherent biases in common data sets impact the performance of algorithms to generalize to situations that humans find easy (e.g., non-standard poses and differences in scale). These biases are present in many existing underwater imagery data sets. In particular the issue of scale and iconic vs. non-iconic views can be severely mismatched between training data and unlabeled data or the data that algorithms are used on for inference. 

\subsection{Hierarchical labeling process}
We have learned that automated image and video annotation should start at a higher taxonomic category. We will be investigating how to (1) use algorithmically generated labels at a higher organizational level and (2) non-expert annotators to accelerate the annotation of large amounts of data to augment \textit{FathomNet} and aid in further algorithm refinement and development. One promising path forward is the use of few-shot learning (Chen, Liu, Kira, Wang, Huang, 2019) to help refine taxonomic labels. In this paradigm, an object detection algorithm trained at one level of hierarchy would create object detections that are filtered to the label that we wish to refine. By using an iterative training and evaluation cycle, we would then rapidly train models for sub-labels of interest. In this way we would be able to rapidly split a hierarchy, even with relatively few labels. These noisy, single-label detectors can then be used in conjunction with the high-performing object detection algorithms to build up a new level of the taxonomy. Once sufficient labels have been produced using this technique, a retraining of the object detection algorithm can be performed, thereby providing a new baseline for object detection.

\subsection{Different types of annotations}
 The main focus of our algorithm efforts on \textit{FathomNet} pertained to single-label image annotations or object-level bounding boxes. There are a variety of different types of annotations that can inform other workflows. One such avenue is multi-label annotation, which would reduce a large amount of the noise associated with training single-label algorithms on multi-label imagery. Very large taxonomic multi-label algorithms are an open area of research. Another annotation type includes segmentation masks, both for instance segmentation of objects (or “things”), as well as semantic segmentation of scenes (or “stuff”). These masks can help characterize benthic scenes for example, or provide more information for recognition algorithms like Mask R-CNN (He, Gkioxari, Dollar,\ \& Girshick, 2017). One of the difficulties in segmentation approaches is generating the training data. This involves drawing appropriate boundaries around every object of interest, which is a task that is even more tedious than drawing bounding boxes. Recent efforts in this area have provided significant speedups for this workflow (Ling, et al, 2019; Acuna, Kar, Fidler, 2019).

\section*{Broader Impacts}

As \textit{FathomNet} continues to develop and incorporate more imagery from other oceanographic community members, we hope that this effort will ultimately enable scientists, explorers, policymakers, storytellers, and the public to better understand how to be stewards of our oceans. There may be unintended consequences in releasing the images and metadata from the seed data set that include illegal poaching or other ocean ecosystem exploits. Therefore it has been a priority for our team throughout this research thrust to take great care in how we craft the metadata released with the images. If successful, FathomNet will aid in the democratization of ocean research and increase accessibility to ocean data and analysis tools, especially in global communities that traditionally do not have ready access to them. Similar to the {\it My Deep Sea, My Backyard} project that originated at the MIT Media Lab Open Ocean Initiative, FathomNet aims to empower communities "around the globe to explore their own deep-sea backyards using low-cost technology, while building lasting in-country capacity."(Amon et al, 2018)

\section*{References}

\small

[1] Barbu, A., Mayo, D., Alverio, J., Luo, W., Wang, C., Gutfreund, D., Tenenbaum, J., and Katz, B. (2019). ObjectNet: A large-scale bias-controlled data set for pushing the limits of object recognition models. {\it Advances in Neural Information Processing Systems 32}, pp. 9453–9463.

[2] Xiao, H., Rasul, K., and Vollgraf, R. (2017). Fashion-MNIST: a Novel Image data set for Benchmarking Machine Learning Algorithms. arXiv:1708.07747.

[3] Y. LeCun, L. Bottou, Y. Bengio, and P. Haffner. (1998) Gradient-based learning applied to document recognition. {\it Proceedings of the IEEE}, 86(11):2278-2324.

[4] Deng, J. et al. (2009). Imagenet: A large-scale hierarchical image database. {\it IEEE conference on computer vision and pattern recognition} pp. 248–255.

[5] Schlining, B.M. and N. Jacobsen Stout (2006). The seed data set’s video annotation and reference system.{\it Proceedings of the Marine Technology Society/Institute of Electrical and Electronics Engineers Oceans Conference} pp. 1-5.

[6] Kirillov, Alexander, Kaiming He, Ross B. Girshick, Carsten Rother and Piotr Dollár. (2019) Panoptic Segmentation. {\it IEEE/CVF Conference on Computer Vision and Pattern Recognition: 9396-9405.}

[7] Stanford, D G and David Nichols Stanford. (2017) Multi-label Classification of Satellite Images with Deep Learning.

[8] Wang, Q., Jia, N., \& Breckon, T.P. (2018). A Baseline for Multi-Label Image Classification Using Ensemble Deep CNN. CoRR, abs/1811.08412.

[9] Li, Y., \& Yeh, M. (2018). Learning Image Conditioned Label Space for Multilabel Classification. CoRR, abs/1802.07460.

[10] Z. Wang, T. Chen, G. Li, R. Xu and L. Lin. (2017) Multi-label Image Recognition by Recurrently Discovering Attentional Regions {\it IEEE International Conference on Computer Vision (ICCV)} pp. 464-472.

[11] Oksuz, Kemal, Baris Can Cam, Sinan Kalkan, and Emre Akbas.(2020) Imbalance Problems in Object Detection: A Review. ArXiv:1909.00169 [Cs]

[12] Gao, M., Li, A., Yu, R., Morariu, V.I., \& Davis, L.S. (2018). C-WSL: Count-Guided Weakly Supervised Localization. {\it ECCV}.

[13] M. Najibi, F. Yang, Q. Wang and R. Piramuthu. (2018) Towards the Success Rate of One: Real-Time Unconstrained Salient Object Detection,{\it IEEE Winter Conference on Applications of Computer Vision (WACV)} pp. 1432-1441.

[14] D. P. Papadopoulos, J. R. R. Uijlings, F. Keller and V. Ferrari. (2016) We Don’t Need No Bounding-Boxes: Training Object Class Detectors Using Only Human Verification. {\it IEEE Conference on Computer Vision and Pattern Recognition (CVPR)} pp. 854-863.

[15] Chattopadhyay, Aditya, Anirban Sarkar, Prantik Howlader and Vineeth N. Balasubramanian. (2018) Grad-CAM++: Generalized Gradient-Based Visual Explanations for Deep Convolutional Networks.{\it IEEE Winter Conference on Applications of Computer Vision} 839-847.

[16] R. Girshick, J. Donahue, T. Darrell and J. Malik. (2014) Rich Feature Hierarchies for Accurate Object Detection and Semantic Segmentation.{\it IEEE Conference on Computer Vision and Pattern Recognition} pp. 580-587.

[17] Chen, W.-Y., Liu, Y.-C., Kira, Z., Wang, Y.-C. F., \& Huang, J.-B. (2019). A Closer Look at Few-shot Classification.{\it International Conference on Learning Representations.}

[18] T. Lin, P. Goyal, R. Girshick, K. He and P. Dollár (2017) Focal Loss for Dense Object Detection.{\it IEEE International Conference on Computer Vision} pp. 2999-3007.

[19] Zhao, Zhong-Qiu, Peng Zheng, Shou-tao Xu, and Xindong Wu.(2019) Object Detection with Deep Learning: A Review. ArXiv:1807.05511

[20] Ovadia, Yaniv, Emily Fertig, Jie Ren, Zachary Nado, D. Sculley, Sebastian Nowozin, Joshua Dillon, Balaji Lakshminarayanan, and Jasper Snoek.(2019) Can You Trust Your Model's Uncertainty? Evaluating Predictive Uncertainty under data set Shift. {\it Advances in Neural Information Processing Systems 32} 13991–14002.

[21] H. Ling, J. Gao, A. Kar, W. Chen, \& S. Fidler (2019). Fast Interactive Object Annotation with Curve-GCN. {\it CoRR} abs/1903.06874.

[22] D. Acuna, A. Kar, \& S. Fidler (2019). Devil is in the Edges: Learning Semantic Boundaries from Noisy Annotations. {\it CoRR} abs/1802.07460.

[23] Amon, Diva, Randi Rotjan, Miriam Simun, Brennan Phillips, Alan Turchik, Katy Croff Bell, Rafael Anta, Kristina Gjerde, and Miriam Simun.(2018) My Deep Sea, My Backyard. {\it Journal of Open Exploration}

\end{document}